\documentclass[conference]{IEEEtran}
\IEEEoverridecommandlockouts
\usepackage{balance}
\usepackage{blindtext}
\usepackage{amsmath,amssymb,amsfonts}
\usepackage{algorithmic}
\usepackage{graphicx}
\usepackage{listings}
\usepackage{booktabs}
\usepackage{textcomp}
\usepackage{xcolor}
\usepackage[binary-units]{siunitx}
\usepackage[numbers,sort&compress]{natbib}
\usepackage{url}
\usepackage{etoolbox}           
\renewcommand{\bfseries}{\fontseries{b}\selectfont} 
\robustify\bfseries             
\newrobustcmd{\B}{\bfseries}    

\newcommand{\funding}{%
  This work was funded by EPSRC grants EP/V052241/1 and EP/S030964/1; and the EU’s Horizon 2020 research and innovation programme under Grant Agreement 945539. Hardware was provided by the Xilinx University Program.
}

\title{FeNN: A RISC-V vector processor for Spiking Neural Network acceleration}
\author{%
  \IEEEauthorblockN{%
    Zainab Aizaz\IEEEauthorrefmark{1}\textsuperscript{\textsection},
    James~C.~Knight\IEEEauthorrefmark{1}\textsuperscript{\textsection},
    and Thomas Nowotny\IEEEauthorrefmark{1}}
  \IEEEauthorblockA{%
    \IEEEauthorrefmark{1}School of Engineering and Informatics,
                    University of Sussex,Brighton, BN1 9QJ, UK \textsuperscript{\textsection} Equal contribution\\
                    Email: \{z.aizaz, j.c.knight, t.nowotny\}@sussex.ac.uk}
  \thanks{\funding}
}

\begin{document}
\maketitle

\begin{abstract}
Spiking Neural Networks~(SNNs) have the potential to drastically reduce the energy requirements of AI systems.
However, mainstream accelerators like GPUs and TPUs are designed for the high arithmetic intensity of standard ANNs so are not well-suited to SNN simulation.
FPGAs are well-suited to applications with low arithmetic intensity as they have high off-chip memory bandwidth and large amounts of on-chip memory.
Here, we present a novel RISC-V-based soft vector processor~(FeNN), tailored to simulating SNNs on FPGAs.
Unlike most dedicated neuromorphic hardware, FeNN is fully programmable and designed to be integrated with applications running on standard computers from the edge to the cloud.
We demonstrate that, by using stochastic rounding and saturation, FeNN can achieve high numerical precision with low hardware utilisation and that a single FeNN core can simulate an SNN classifier faster than both an embedded GPU and the Loihi neuromorphic system.

\end{abstract}

\begin{IEEEkeywords}
SNN, FPGA, RISC-V
\end{IEEEkeywords}

\section{Introduction}
Artificial Neural Networks~(ANNs) have demonstrated super-human performance in areas ranging from image classification to playing go~\citep{silver_mastering_2016}.
However, training and using ANNs is computationally very expensive and only possible due to modern hardware and software.
The primary computational primitive ANNs require is matrix multiplication, which has a very high arithmetic intensity.
Numerous accelerators including GPUs and TPUs have been designed to accelerate matrix multiplication, typically by streaming data from high bandwidth external memory into systolic arrays of multiply-accumulate units.
However, high arithmetic intensity leads to high energy requirements, representing a growing problem for the AI industry.
At the same time, the \emph{biological} neural network of the human brain can still outperform ANNs in many tasks while only consuming \SI{20}{\watt}.

While ANNs are \emph{inspired} by the brain, they differ in several key areas, one of which is that
biological neurons communicate with sparse binary events called \emph{spikes}.
This has been incorporated in artificial Spiking Neural Networks~(SNNs) which have already been successful in spatio-temporal processing applications such as audio denoising and autonomous driving~\citep{shrestha_efficient_2024}.
As spikes carry no `value' to multiply, simulating SNNs does not require matrix multiplication and thus has a much lower arithmetic intensity. 
\citet{bautembach_even_2021,Knight2022} previously showed that SNNs can be simulated effectively on GPUs.
However, the low arithmetic intensity of SNNs means that SNN simulations can easily saturate a GPU's memory bandwidth but leave its compute resources under-utilised.
Furthermore, while low-precision weights have been widely adopted in GPU designs to reduce memory bandwidth requirements, they are typically handled by the systolic MAC arrays which do not support the operations required for simulating SNNs at low precision~\citep{fernandez-hart_posit_2024,Hopkins2020}.
Finally, because simulating SNNs is an iterative process and GPUs have limited capacity for global synchronisation, several kernels typically are launched every simulation timestep causing a significant overhead (order of \SI{10}{\micro\second}).

Digital neuromorphic systems~\citep{Davies2018,gonzalez_spinnaker2_2024} are tailored to the low arithmetic intensity of SNN simulations, typically only having modest computational resources but large amounts of fast local memory.
These systems are very efficient but typically not flexible enough for developing \emph{new} algorithms.
Furthermore, neuromorphic systems are not available off the shelf and unlike GPUs cannot easily be integrated with applications running on standard computers as they are typically standalone systems.
Field-Programmable Gate Arrays~(FPGAs) represent an interesting middle ground as they can be integrated seamlessly with standard computers and not only offer high \emph{external} memory bandwidth similar to GPUs but also large amounts of \emph{on-chip memory} that can be accessed in a single cycle~(around \SI{40}{\mebi\byte} on a high-end Alveo U55C).
Although FPGAs also have DSP cores to enable single-cycle integer multiplication, they run at relatively low clock speeds (typically a few \SI{100}{\mega\hertz}) so cannot compete with GPUs for arithmetically-intensive tasks.
This has meant that they have thus far only been used for inference with standard ANNs~\citep{nechi_fpga-based_2023}.
However, FPGAs are well-suited to SNN simulations because of their lower arithmetic intensity.
Over the years, there has been a plethora of fixed-function FPGA SNN accelerators which demonstrate this~\citep{Moore2012,kauth_neuroaix-framework_2023}.
However, such systems have a high entry barrier as they require re-synthesis using FPGA tools and potentially even writing HDL if the emulated model changes.
\citet{Naylor2013} demonstrated that a fully programmable vector soft-core processor running on an Altera Stratix IV FPGA could simulate SNNs faster than the then state-of-the-art, and more recent programmable systems~\citep{chen_gaban_2022,sripad_snavareal-time_2018} have shown that this approach continues to be competitive on newer FPGA architectures like AMD's UltraScale+.

Here, we present our FPGA-Enhanced Neural Network~(FeNN) accelerator: a vector soft-processor, tailored to the needs of SNN simulation and the hardware resources available on modern FPGAs.
Uniquely, it is designed to operate as an in-memory accelerator, capable of scaling from embedded systems based on Zynq MPSoCs to workstations or cloud nodes equipped with large accelerators such as the Xilinx Alveo U55C.
We show that, by using stochastic rounding and saturation, complex spiking neuron models can be simulated accurately using \SI{16}{\bit} fixed-point arithmetic and demonstrate a recurrent SNN that classifies spoken digits~\citep{Cramer2020} on FeNN.
This classifier runs significantly faster than a similar model running on Loihi and twice as fast as the same model running on an embedded GPU while using half the energy.

\begin{figure}
    \centering
    \includegraphics[height=45mm]{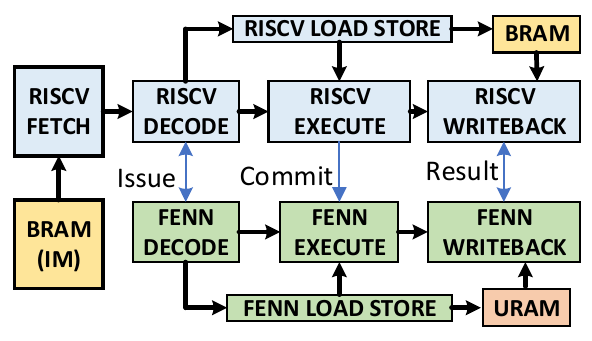}
    \caption{Block diagram of FeNN, tightly coupled to CV32E40X RISC-V core. Black arrows indicate pipeline stages and key signals. Blue arrows highlight the `Issue', `Commit' and `Result' signals defined by the CV32E40X's XIF extension interface.}
    \label{fig:fenn}
\end{figure}
\begin{figure*}
    \includegraphics{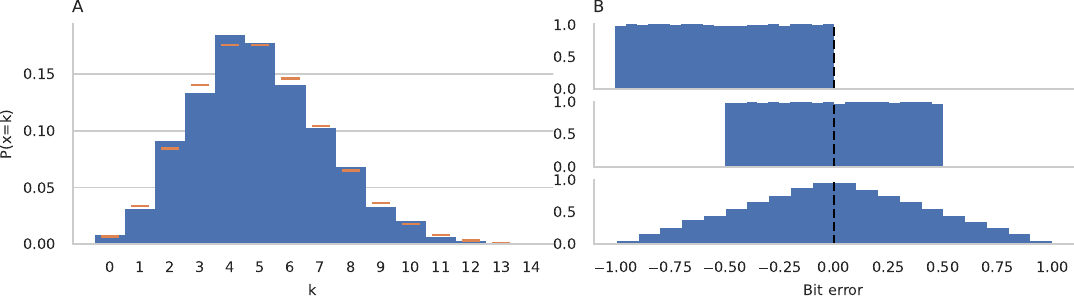}
    \caption{\textbf{(A)} Histogram of \num{3200} Poisson variates sampled from distribution with $\lambda = 5$, generated on FeNN. Horizontal lines indicate Probability Mass Function of distribution.
    \textbf{(B)} Histograms showing the bit error distribution of \num{21760} random S0.15 multiplication operations with (top) round-to-zero (middle) round-to-nearest and (bottom) stochastic rounding (after \citet{Hopkins2020}).}
    \label{fig:random}
\end{figure*}
\section{Methods}
\subsection{Scalar core}
Soft vector processors can either be \emph{tightly coupled}, where individual instructions are offloaded from the scalar processor to the vector processor or \emph{decoupled}, where `kernels' of vector instructions are assembled separately and copied to the accelerator.
While tightly coupled architectures allow fine-grained communication between scalar and vector cores, their performance can be restricted by the throughput of the scalar processor and each application must be compiled for the novel architecture~\citep{Naylor2013}.
However, decoupled vector processors have to implement the entire control flow for the desired algorithms which would introduce significant complexity.

We have developed a best-of-both-worlds architecture where our FeNN vector processor works as a decoupled, in-memory accelerator for a standard host CPU (a quad-core ARM Cortex-A53 on the Kria KV260) running Linux but is tightly coupled to its own scalar core.
The scalar core is a CV32E40X developed by the Open Hardware Group and is a small and efficient, 32-bit, in-order RISC-V core with a 4-stage pipeline.
Data and control hazards notwithstanding, this core can issue 1 instruction per cycle and provides a convenient extension interface upon which to implement our co-processor.
In order to make the CV32E40X more suitable for FPGA synthesis, we have simplified its fetch and load-store stages to work with synchronous Block RAMS -- removing the need for any instruction prefetch or alignment logic.
\begin{table}
\caption{Hardware utilisation. Bracketed values are per lane/PE.}
  \centering
  \begin{tabular}{r *{4}{S[detect-weight,   
           mode=text,table-format=1.0]}}
    \toprule
        {Design}        & {Lanes} & {Frequency} & \multicolumn{2}{S}{Utilisation}\\
                        &         & {[\si{\mega\hertz}]} & {LUT} & {FF} \\
    \midrule
        SNAVA~\citep{sripad_snavareal-time_2018} & 100         & 125                          & {\num{148774} (\num{1488})} & {\num{97824} (\num{978})}\\
        GaBAN~\citep{chen_gaban_2022} & 8           & 286                          & {\num{28759} (\num{3595})} & {\num{9605} (\num{1201})} \\
        \textbf{FeNN}   & \B 32          & \B 166                          & \textbf{\num{32915} (\num{1028})} & \textbf{\num{32547} (\num{1017})} \\
  \end{tabular}
  \label{tab:hardware}
\end{table}
\subsection{3-stage pipelined vector co-processor}
While it is common for inference accelerators to use \SI{8}{\bit} or even \SI{4}{\bit} weights, we want to offer more flexibility and, in future, also accelerate SNN training workloads. Hence, our accelerator operates on \SI{16}{\bit} types throughout.
Specifically, FeNN operates on vectors of 32 \SI{16}{\bit} values using a SIMD architecture and a standard register-register instruction set, backed by a three-port vector register file with 32 \SI{512}{\bit} registers, implemented using distributed memory.
The majority of on-chip memory on UltraScale+ FPGAs is provided through `wide'~(\SI{72}{\bit}) `UltraRAM' memory which we use to implement local memories, large enough to store all model state in our current prototype.
These memories are created using 8 parallel sets of chained UltraRAM blocks, enabling FeNN to load and store entire vectors in a single cycle.

In order to provide scope for further expansion, FeNN is implemented within a full \SI{30}{\bit} RISC-V instruction set quadrant (binary prefix \lstinline{10}).
Within this quadrant, our initial FeNN design implements a minimal instruction set supporting: 
\begin{itemize}
    \item Addition, subtraction and multiplication of vectors.
    \item Loading and storing of vectors in UltraRAM.
    \item Moving data between scalar and vector registers.
    \item Generating random numbers.
    \item Computing masks by comparing vectors.
    \item Selecting between elements of two vectors using masks.
\end{itemize}
Multiplication is implemented using DSP blocks and, as these can operate at several times our operating frequency, this leaves time in the execute cycle to apply a \SI{16}{\bit} barrel shift to the result in order to implement fixed-point multiplication.
Because FeNN has 32 vector lanes, the masks used for non-uniform control flow conveniently fit in a standard 32-bit RISC-V register.
Thus, in our tightly-coupled architecture, masks can be processed using vector and scalar instructions, removing hardware requirements for separate mask registers.

\subsubsection{Pipeline}
Figure~\ref{fig:fenn} illustrates how FeNN is pipelined alongside the CV32E40X with synchronisation occurring through the XIF extension interface.
All instructions can be executed within a single execute clock cycle and we include additional logic between the writeback and execute stages to enable bypassing of RAW hazards between vector registers.
UltraRAM resources are synchronous with a 1-cycle latency so addresses are calculated in the execute cycle and values read from memory are written back to the register file in the writeback cycle.

\subsubsection{Random number generation}
\label{sec:meth_random}
Efficient pseudorandom number generation  is important in many SNN models as well as for stochastic rounding~\citep{Hopkins2020} so an efficient Random Number Generator~(RNG) is a key requirement for FeNN.
Because FeNN operates on \SI{16}{\bit} numbers, we require an RNG with a small state size that can generate \SI{16}{\bit} of randomness per cycle in each vector lane.
Inspired by its use in the Propeller 2 microcontroller, we chose a Xoroshiro32++ generator~\citep{blackman_scrambled_2021} which is relatively high quality and extremely hardware friendly -- requiring only a handful of addition, XOR, rotation and shift operations.
However, if it were to be implemented as a standard RISC-V instruction -- which reads at most two operands and writes back another -- the instruction to generate a random number would need to read \SI{32}{\bit} and write \SI{48}{\bit} of state per cycle requiring extra register file ports.
Furthermore, incorporating the RNG into a stochastic multiplication instruction would require reading two more operands and writing back yet another.
Instead, we add two special two-port registers dedicated to holding the RNG state.
These can be populated from vector memory using special variants of the vector load instruction.

\subsubsection{Stochastic rounding}
\label{sec:meth_stoc}
A standard fixed point multiplication is performed by calculating \lstinline{(A * B) >> N} where \lstinline{A} and \lstinline{B} are two fixed-point encoded operands and \lstinline{N} is the number of fractional bits in the representation.
The right shift by \lstinline{N} truncates the result leading to round-to-zero behaviour.
However, \citet{mikaitis_stochastic_2021} demonstrated that rounding can be implemented simply by turning the multiplication into a MAC (\lstinline{((A * B) + R) >> N}) which is supported in hardware by the DSP blocks.
Round-to-nearest can be implemented simply by setting \lstinline{R} to 0.5 (in fixed-point with \lstinline{N} fractional bits) and stochastic rounding by adding \lstinline{N} bits of randomness (the \emph{mean} of which will be 0.5) from the RNG.

\subsection{System-on-Chip design}
In our current prototype we have instantiated a single FeNN core on an AMD Kria KV260 development board.
The data and instruction BRAMs connected to the scalar RISC-V core are exposed to the ARM core via standard AXI BRAM controllers.
Similarly, control and status registers are exposed via an AXI GPIO controller.
These resources are then memory mapped by user-space applications running on the ARM core.
Applications use a C++-based Just-in-Time assembler to generate FeNN instructions and copy them, alongside any initial state, to the memory-mapped BRAMs.
FeNN is then reset and simulates multiple SNN timesteps while the host polls, waiting for FeNN to finish.

\subsection{Adaptive Leaky Integrate-and-Fire neuron model}
\label{sec:methods/alif}
Standard Recurrent SNNs~(RSNNs) using Leaky Integrate-and-Fire~(LIF) neurons have inferior short-term memory capacities compared to recurrent ANN architectures such as LSTMs.
Adaptive LIF~(ALIF) neurons -- which augment a standard LIF neuron with a slowly adapting threshold -- are one solution to this problem~\citep{Bellec2018}.
The dynamics of an ALIF neuron's membrane voltage $V$  can be calculated using an Exponential Euler scheme,
\begin{align}
    V[t + 1] =& \alpha V[t] + I^\text{syn}[t + 1] - S[t] V_\text{th}
\end{align}
where $\alpha = e^\frac{-1}{\tau_m}$ controls the neuron's leak ($\tau_m = 20$), $I^\text{syn}$ is the input current and $V_\text{th} = 0.6$ is the baseline firing threshold which is subtracted from $V$ when a spike occurs.
Spikes are triggered when the membrane voltage crosses the adaptive threshold:
\begin{align}
    S[t] =& H\left(V[t] - (V_\text{th} + \beta A[t])\right) 
\end{align}
where $\beta=0.0174$ is the adaptation strength and $H$ the Heaviside function.
The adaptation $A$ evolves as
\begin{align}
    A[t + 1] =& \rho A[t] + S[t]
\end{align}
where $\rho = e^\frac{-1}{\tau_a}$ controls the rate of adaptation ($\tau_a = 2000$).
\vfill
\begin{figure*}
    \includegraphics{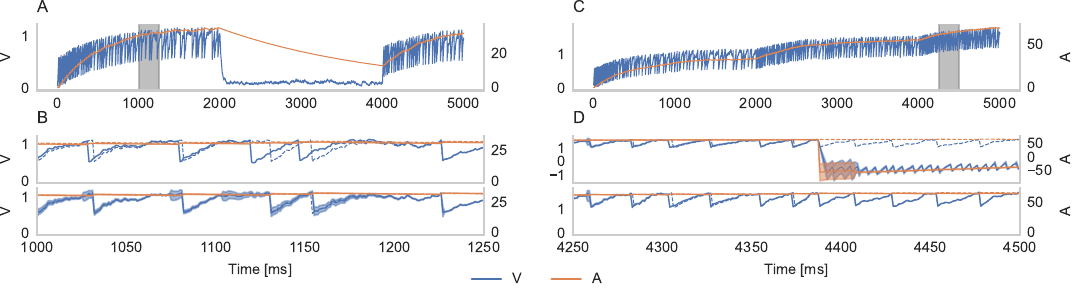}
    \caption{ALIF neuron simulations \textbf{(A)} ALIF stimulated with two periods of Poisson spiking input, separated by a period of low background rate, simulated using \SI{64}{\bit} floating point \textbf{(B)}~Detail from the shaded area in A) comparing \SI{16}{\bit} fixed point (solid lines) against \SI{64}{\bit} floating point (dashed lines) using standard round-to-zero (top) and stochastic rounding (bottom).
    \textbf{(C)} ALIF stimulated by Poisson spiking input with increasing rates, simulated using \SI{64}{\bit} floating point \textbf{(D)}~Detail from C) comparing \SI{16}{\bit} fixed point (solid lines) against \SI{64}{\bit} floating point (dashed lines) using stochastic rounding (top) and stochastic rounding with saturation (bottom).
    Simulations with stochastic rounding were run 32 times and the standard deviations are shown as shaded areas.}
    \label{fig:alif_rounding}
\end{figure*}
\section{Results}
All code used to generate the results in this section and a bitstream for programming a Kria KV260 with FeNN are available at \url{https://github.com/neworderofjamie/riscv_ise}.
\subsection{Synthesis}
We synthesised our design using Xilinx Vivado 2023.2.
Table~\ref{tab:hardware} compares hardware utilisation and clock speed with recent programmable FPGA-based SNN accelerators~\citep{sripad_snavareal-time_2018,chen_gaban_2022}.
Due to our focus on \SI{16}{\bit} fixed point, a FeNN core requires significantly fewer LUTs per-lane than other designs.
On the other hand, FeNN's deeper pipeline means that it requires more FlipFlops than the other designs, but these are not the scarcest resource and its deeper pipeline means FeNN can process neurons in fewer clock cycles (approximately 20 compared to 85 for GaBAN).

\subsection{Random number generation}
In order to test the random number generator described in Section~\ref{sec:meth_random}, we implemented a Poisson generator on FeNN using the direct method first described by~\citet{knuth2014art} and sampled \num{3200} variates with $\lambda=5$.
As Poisson processes are often used to encode stimuli in SNN models~\citep{pfeiffer_deep_2018}, this is a representative test of how random number generation is likely to be used on FeNN.
Figure~\ref{fig:random}A shows that the sampled variates follow the desired distribution and our timing measurements show that, on average, the Poisson generator produces \num{32} variates every \num{81} clock cycles (corresponding to around \SI{64E6}{\per\second}).

\subsection{Rounding and saturation}
In order to test the correctness of the rounding mechanisms described in Section~\ref{sec:meth_stoc}, we repeated the experiment described by \citet{Hopkins2020} and performed \num{21760} multiplication operations with round-to-zero, round-to-nearest and stochastic rounding on random \SI{16}{\bit} operands in S0.15 fixed-point format (S0.15 can only represent numbers with absolute value $<1$ so no overflow will occur) using FeNN.
Figure~\ref{fig:random}B shows that the bit error distributions obtained by comparing the results to \SI{64}{\bit} floating point multiplication match those obtained by \citeauthor{Hopkins2020}.

Using ALIF neurons has been shown to significantly improve the performance of SNN models~\citep{Bellec2018}.
However, accurately simulating their dynamics can prove challenging on digital hardware as fixed-point numbers implemented with standard round-to-zero integer operations prematurely underflow to zero.
Figure~\ref{fig:alif_rounding} shows simulations of ALIF neurons driven by challenging input spike trains. Figure~\ref{fig:alif_rounding}B highlights a period where the \SI{16}{\bit} fixed point simulation with round-to-zero diverges from the \SI{64}{\bit} floating point reference simulation while the version with stochastic rounding follows closely.
Overall, stochastic rounding significantly reduces the Normalised Root-Mean Square Error~(NRMSE) compared to the floating point reference (NRMSE of V reduces from \num{0.13} to \num{0.045} and A from \num{0.019} to {0.0066}).

When training SNN models or performing inference on live data, stochastic rounding alone is not sufficient for numerically safe simulations.
As figure~\ref{fig:alif_rounding}D shows, if a neuron's input increases temporarily above the calibrated range of the fixed-point formats, state variables overflow and wrap.
This results in catastrophic failure (NRMSE of V of \num{0.29} and A of \num{0.39}) that can, however, be mitigated by saturating additions and subtractions (NRMSE of V of \num{0.098} and A of \num{0.019}).
\begin{figure*}
    \includegraphics{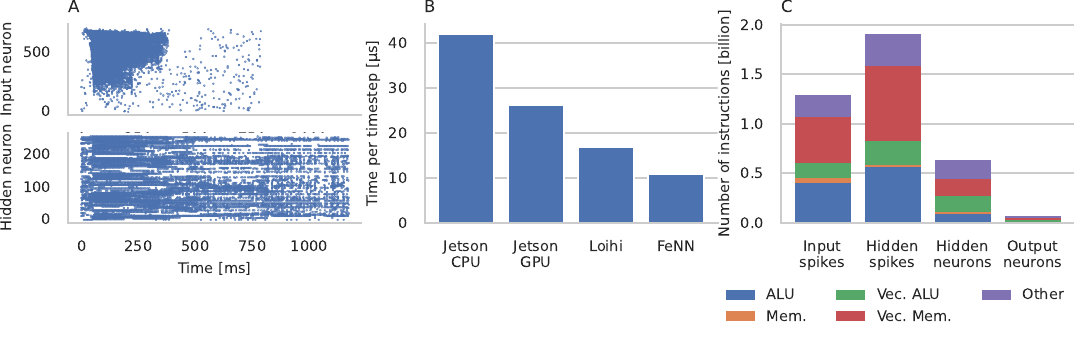}
    \caption{SHD inference running on FeNN \textbf{(A)} Raster plot of input and hidden neuron activity during one example. \textbf{(B)} Comparison of average time (over the whole test set) taken to simulate a single SNN timestep on FeNN with Jetson Orin Nano using GeNN (batch size=1)~\citep{Knight2018} and similar RSNN running on Loihi~\citep{rao_long_2022}. GPU times were measured using CUDA timing events and CPU and FeNN times using \lstinline{std::chrono::high_resolution_clock} \textbf{(C)}~Distribution of different types of instructions in different stages of the simulation.}
    \label{fig:shd_perf}
\end{figure*}

\subsection{Classification}
To demonstrate the performance of our FeNN prototype on a real machine learning task, we trained an ALIF-based Recurrent SNN~(RSNN) classifier with a single \num{256} neuron hidden layer on the Spiking Heidelberg Digits~(SHD) dataset~\cite{Cramer2020}, quantized the trained weights and deployed it onto FeNN.
Running on FeNN, this classifier obtains an accuracy of \SI{79.5(0.44)}{\percent} on the test set -- an insignificant difference in performance compared to the \SI{32}{\bit} floating point GeNN implementation which obtains \SI{79.6}{\percent}.
Figure~\ref{fig:shd_perf}A shows the input spikes from an example digit and the hidden layer activity within the network.
In Figure~\ref{fig:shd_perf}B, we compare the average time (over the whole SHD test set) taken to simulate a single SNN timestep using FeNN and a Jetson Orin Nano (as this has a similar form factor and power usage to a Kria KV 260) using GeNN~\citep{Knight2018}.
Additionally, we include the time reported by \citet{rao_long_2022} for a similar ALIF-based RSNN classifier on the Loihi~\citep{Davies2018} neuromorphic system.
\begin{table}
\caption{SHD classification energy. `Total inference energy' is all energy spent during inference. Idle power and time spent copying data are removed to calculate `Simulation energy' and this is divided by the total number of synaptic events to get `Energy per synaptic event'.}
  \centering
  \begin{tabular}{r *{3}{S[detect-weight,   
           mode=text,table-format=1.0]}}
    \toprule
        {Device}    & {Total inference}        & {Simulation}          & {Energy per synaptic}\\
                    & {energy [\si{\joule}]}    & {energy [\si{\joule}]}& {event [\si{\nano\joule}]} \\
    \midrule
        \textbf{FeNN}      & \B 347                       & \B 44                & \B 8 \\
        Jetson GPU  & 645                       & 97                & 18 \\
        Jetson CPU  & 956                       & 113               & 20 \\
  \end{tabular}
  \label{tab:energy}
\end{table}

\subsection{Energy usage}
Neither the Jetson Orin Nano or Kria KV260 provide a sensor for measuring the power used by the GPU or FPGA accelerator.
Therefore, we recorded the power usage during SHD inference using a consumer power measurement device at the mains socket, captured the output with a webcam and performed optical character recognition using the \lstinline{ssocr} Ubuntu package.
We then integrated the power time series using the trapezoidal rule to calculate the energy spent during the whole simulation (we dis-regarded initialisation as it was implemented very differently in the two simulations).
As Table~\ref{tab:energy} shows, FeNN takes approximately half the energy of the Jetson GPU and a third of the Jetson CPU to evaluate the SHD test set.
However, both systems were running desktop Linux operating systems so the majority of this energy goes to driving the display, performing background tasks etc.
Furthermore, a significant amount of time is spent streaming SHD digits to the accelerators which is not currently well-optimised on either system.
As neither of these factors are of direct interest to this work, we also show simulation energy with these removed and energy per-synaptic event in Table~\ref{tab:energy}.
Again, these illustrate that FeNN takes approximately half the energy of the Jetson for the same task.

\section{Conclusions and Future Work}
Here we have presented a first prototype of our FeNN accelerator which is already capable of efficiently simulating a moderately-sized RSNN classifier, faster than both embedded GPU and neuromorphic systems.
Furthermore, FeNN performs these simulations using around half of the energy of an embedded GPU.

Figure~\ref{fig:shd_perf}C shows the distribution of different instruction types executed by FeNN while evaluating the SHD test.
The ratio of `Vector Memory'  and `Vector ALU' instructions clearly illustrates the low arithmetic intensity of SNN simulation, especially in the spike-processing code where this is 3:1.
When processing spikes, a significant number of scalar ALU instructions are spent iterating over spikes and calculating the addresses to load weights from.
To reduce this overhead and also allow FeNN to simulate larger models with weights stored in external memory, we plan to develop Direct Memory Access (DMA) hardware which which will calculate addresses from spikes and asynchronously copy weights from external memory to UltraRAM.
This DMA hardware will also enable the transfer of data between FeNN cores using an AXI crossbar.
As well as these system-on-chip features, we also plan on adding lane-local memories -- for efficient processing of sparse connectivity~\citep{Naylor2013} -- and instructions to perform reductions across vector lanes to each core.
Finally, the SNNs simulated here have all been hand-coded in RISC-V assembly language.
However, we are developing a backend for our GeNN library~\citep{Knight2018} to convert high-level model descriptions into FeNN instructions.

With these developments in place, we intend to scale up our prototype to a \num{32} core version instantiated on an Alveo U55C accelerator which will be capable of batched Eventprop~\citep{wunderlich_event-based_2021} training in addition to inference.
Because Eventprop training has a similar computational complexity as inference, we would expect this system to be able to \emph{train} models at a similar speed and energy cost to the inference task we perform here.
For \emph{inference}, the \SI{8}{\nano\joule} per-SOP achieved by FeNN is not competitive with the latest neuromorphic systems (at least partly due to the $10\times$ overhead of FPGA vs ASIC logic~\citep{Kuon2006}).
However, due to its greater flexibility, including the ability to support inference and Eventprop training in future, FeNN is a valuable intermediate step towards the wider adoption of neuromorphic technologies.
\balance 
\bibliography{fpga}

\begin{thebibliography}{24}
\providecommand{\natexlab}[1]{#1}
\providecommand{\url}[1]{\texttt{#1}}
\expandafter\ifx\csname urlstyle\endcsname\relax
  \providecommand{\doi}[1]{doi: #1}\else
  \providecommand{\doi}{doi: \begingroup \urlstyle{rm}\Url}\fi

\bibitem[Silver et~al.(2016)Silver, Huang, Maddison, Guez, Sifre, van~den
  Driessche, Schrittwieser, Antonoglou, Panneershelvam, Lanctot, Dieleman,
  Grewe, Nham, Kalchbrenner, Sutskever, Lillicrap, Leach, Kavukcuoglu, Graepel,
  and Hassabis]{silver_mastering_2016}
David Silver, Aja Huang, Chris~J. Maddison, Arthur Guez, Laurent Sifre, George
  van~den Driessche, Julian Schrittwieser, Ioannis Antonoglou, Veda
  Panneershelvam, Marc Lanctot, Sander Dieleman, Dominik Grewe, John Nham, Nal
  Kalchbrenner, Ilya Sutskever, Timothy Lillicrap, Madeleine Leach, Koray
  Kavukcuoglu, Thore Graepel, and Demis Hassabis.
\newblock Mastering the game of {Go} with deep neural networks and tree search.
\newblock \emph{Nature}, 529\penalty0 (7587):\penalty0 484--489, 2016.
\newblock ISSN 0028-0836.
\newblock \doi{10.1038/nature16961}.
\newblock URL \url{http://www.nature.com/doifinder/10.1038/nature16961}.

\bibitem[Shrestha et~al.(2024)Shrestha, Timcheck, Frady, Campos-Macias, and
  Davies]{shrestha_efficient_2024}
Sumit~Bam Shrestha, Jonathan Timcheck, Paxon Frady, Leobardo Campos-Macias, and
  Mike Davies.
\newblock Efficient {Video} and {Audio} {Processing} with {Loihi} 2.
\newblock In \emph{{ICASSP} 2024 - 2024 {IEEE} {International} {Conference} on
  {Acoustics}, {Speech} and {Signal} {Processing} ({ICASSP})}, pages
  13481--13485, Seoul, Korea, Republic of, April 2024. IEEE.
\newblock ISBN 9798350344851.
\newblock \doi{10.1109/ICASSP48485.2024.10448003}.
\newblock URL \url{https://ieeexplore.ieee.org/document/10448003/}.

\bibitem[Bautembach et~al.(2021)Bautembach, Oikonomidis, and
  Argyros]{bautembach_even_2021}
Dennis Bautembach, Iason Oikonomidis, and Antonis Argyros.
\newblock Even {Faster} {SNN} {Simulation} with {Lazy}+{Event}-driven
  {Plasticity} and {Shared} {Atomics}.
\newblock In \emph{2021 {IEEE} {High} {Performance} {Extreme} {Computing}
  {Conference} ({HPEC})}, pages 1--8, September 2021.
\newblock \doi{10.1109/HPEC49654.2021.9622805}.
\newblock URL
  \url{https://ieeexplore.ieee.org/document/9622805/?arnumber=9622805}.
\newblock ISSN: 2643-1971.

\bibitem[Knight and Nowotny(2022)]{Knight2022}
James~C Knight and Thomas Nowotny.
\newblock Efficient {GPU} training of {LSNNs} using {eProp}.
\newblock In \emph{Neuro-{Inspired} {Computational} {Elements} {Conference}},
  pages 8--10, New York, NY, USA, March 2022. ACM.
\newblock ISBN 978-1-4503-9559-5.
\newblock \doi{10.1145/3517343.3517346}.
\newblock URL \url{https://dl.acm.org/doi/10.1145/3517343.3517346}.

\bibitem[Fernandez-Hart et~al.(2024)Fernandez-Hart, Knight, and
  Kalganova]{fernandez-hart_posit_2024}
T.~Fernandez-Hart, James~C. Knight, and T.~Kalganova.
\newblock Posit and floating-point based {Izhikevich} neuron: {A} {Comparison}
  of arithmetic.
\newblock \emph{Neurocomputing}, page 127903, May 2024.
\newblock ISSN 09252312.
\newblock \doi{10.1016/j.neucom.2024.127903}.
\newblock URL
  \url{https://linkinghub.elsevier.com/retrieve/pii/S092523122400674X}.

\bibitem[Hopkins et~al.(2020)Hopkins, Mikaitis, Lester, and
  Furber]{Hopkins2020}
Michael Hopkins, Mantas Mikaitis, Dave~R Lester, and Steve Furber.
\newblock Stochastic rounding and reduced-precision fixed-point arithmetic for
  solving neural ordinary differential equations.
\newblock \emph{Philosophical Transactions of the Royal Society A:
  Mathematical, Physical and Engineering Sciences}, 378\penalty0
  (2166):\penalty0 20190052, March 2020.
\newblock ISSN 1364-503X.
\newblock \doi{10.1098/rsta.2019.0052}.
\newblock URL
  \url{https://royalsocietypublishing.org/doi/10.1098/rsta.2019.0052}.

\bibitem[Dav(2018)]{Davies2018}
Loihi : a {Neuromorphic} {Manycore} {Processor} with {On}-{Chip} {Learning}.
\newblock \emph{IEEE Micro}, 30\penalty0 (1):\penalty0 82--99, 2018.
\newblock \doi{10.1109/MM.2018.112130359}.

\bibitem[Gonzalez et~al.(2024)Gonzalez, Huang, Kelber, Nazeer, Langer, Liu,
  Lohrmann, Rostami, Schöne, Vogginger, Wunderlich, Yan, Akl, and
  Mayr]{gonzalez_spinnaker2_2024}
Hector~A. Gonzalez, Jiaxin Huang, Florian Kelber, Khaleelulla~Khan Nazeer, Tim
  Langer, Chen Liu, Matthias Lohrmann, Amirhossein Rostami, Mark Schöne,
  Bernhard Vogginger, Timo~C. Wunderlich, Yexin Yan, Mahmoud Akl, and Christian
  Mayr.
\newblock {SpiNNaker2}: {A} {Large}-{Scale} {Neuromorphic} {System} for
  {Event}-{Based} and {Asynchronous} {Machine} {Learning}, January 2024.
\newblock URL \url{http://arxiv.org/abs/2401.04491}.
\newblock arXiv:2401.04491 [cs].

\bibitem[Nechi et~al.(2023)Nechi, Groth, Mulhem, Merchant, Buchty, and
  Berekovic]{nechi_fpga-based_2023}
Anouar Nechi, Lukas Groth, Saleh Mulhem, Farhad Merchant, Rainer Buchty, and
  Mladen Berekovic.
\newblock {FPGA}-based {Deep} {Learning} {Inference} {Accelerators}: {Where}
  {Are} {We} {Standing}?
\newblock \emph{ACM Transactions on Reconfigurable Technology and Systems},
  16\penalty0 (4):\penalty0 1--32, December 2023.
\newblock ISSN 1936-7406, 1936-7414.
\newblock \doi{10.1145/3613963}.
\newblock URL \url{https://dl.acm.org/doi/10.1145/3613963}.

\bibitem[Moore et~al.(2012)Moore, Fox, Marsh, Markettos, and
  Mujumdar]{Moore2012}
Simon~W. Moore, Paul~J. Fox, Steven~J.T. Marsh, a.~Theodore Markettos, and Alan
  Mujumdar.
\newblock Bluehive - {A} {Field}-{Programable} {Custom} {Computing} {Machine}
  for {Extreme}-{Scale} {Real}-{Time} {Neural} {Network} {Simulation}.
\newblock In \emph{2012 {IEEE} 20th {International} {Symposium} on
  {Field}-{Programmable} {Custom} {Computing} {Machines}}, pages 133--140,
  Toronto, April 2012. IEEE.
\newblock ISBN 978-1-4673-1605-7.
\newblock \doi{10.1109/FCCM.2012.32}.
\newblock URL
  \url{http://ieeexplore.ieee.org/lpdocs/epic03/wrapper.htm?arnumber=6239804}.

\bibitem[Kauth et~al.(2023)Kauth, Stadtmann, Sobhani, and
  Gemmeke]{kauth_neuroaix-framework_2023}
Kevin Kauth, Tim Stadtmann, Vida Sobhani, and Tobias Gemmeke.
\newblock {neuroAIx}-{Framework}: design of future neuroscience simulation
  systems exhibiting execution of the cortical microcircuit model 20× faster
  than biological real-time.
\newblock \emph{Frontiers in Computational Neuroscience}, 17:\penalty0 1144143,
  April 2023.
\newblock ISSN 1662-5188.
\newblock \doi{10.3389/fncom.2023.1144143}.
\newblock URL
  \url{https://www.frontiersin.org/articles/10.3389/fncom.2023.1144143/full}.

\bibitem[Naylor et~al.(2013)Naylor, Fox, Markettos, and Moore]{Naylor2013}
Matthew Naylor, Paul~J. Fox, A.~Theodore Markettos, and Simon~W. Moore.
\newblock Managing the {FPGA} memory wall: {Custom} computing or vector
  processing?
\newblock In \emph{2013 23rd {International} {Conference} on {Field}
  programmable {Logic} and {Applications}}, pages 1--6, Porto, September 2013.
  IEEE.
\newblock ISBN 978-1-4799-0004-6.
\newblock \doi{10.1109/FPL.2013.6645538}.
\newblock URL \url{http://ieeexplore.ieee.org/document/6645538/}.

\bibitem[Chen et~al.(2022)Chen, Yang, and Zhang]{chen_gaban_2022}
Jiajie Chen, Le~Yang, and Youhui Zhang.
\newblock {GaBAN}: a generic and flexibly programmable vector neuro-processor
  on {FPGA}.
\newblock In \emph{Proceedings of the 59th {ACM}/{IEEE} {Design} {Automation}
  {Conference}}, pages 931--936, San Francisco California, July 2022. ACM.
\newblock ISBN 978-1-4503-9142-9.
\newblock URL \url{https://dl.acm.org/doi/10.1145/3489517.3530561}.

\bibitem[Sripad et~al.(2018)Sripad, Sanchez, Zapata, Pirrone, Dorta, Cambria,
  Marti, Krishnamourthy, and Madrenas]{sripad_snavareal-time_2018}
Athul Sripad, Giovanny Sanchez, Mireya Zapata, Vito Pirrone, Taho Dorta,
  Salvatore Cambria, Albert Marti, Karthikeyan Krishnamourthy, and Jordi
  Madrenas.
\newblock {SNAVA}—{A} real-time multi-{FPGA} multi-model spiking neural
  network simulation architecture.
\newblock \emph{Neural Networks}, 97:\penalty0 28--45, January 2018.
\newblock ISSN 08936080.
\newblock \doi{10.1016/j.neunet.2017.09.011}.
\newblock URL
  \url{https://linkinghub.elsevier.com/retrieve/pii/S0893608017302150}.

\bibitem[Cramer et~al.(2022)Cramer, Stradmann, Schemmel, and Zenke]{Cramer2020}
Benjamin Cramer, Yannik Stradmann, Johannes Schemmel, and Friedemann Zenke.
\newblock The {Heidelberg} {Spiking} {Data} {Sets} for the {Systematic}
  {Evaluation} of {Spiking} {Neural} {Networks}.
\newblock \emph{IEEE Transactions on Neural Networks and Learning Systems},
  pages 1--14, 2022.
\newblock ISSN 2162-237X.
\newblock \doi{10.1109/TNNLS.2020.3044364}.
\newblock URL \url{https://ieeexplore.ieee.org/document/9311226/}.
\newblock arXiv: 1910.07407.

\bibitem[Blackman and Vigna(2021)]{blackman_scrambled_2021}
David Blackman and Sebastiano Vigna.
\newblock Scrambled {Linear} {Pseudorandom} {Number} {Generators}.
\newblock \emph{ACM Transactions on Mathematical Software}, 47\penalty0
  (4):\penalty0 1--32, December 2021.
\newblock ISSN 0098-3500, 1557-7295.
\newblock \doi{10.1145/3460772}.
\newblock URL \url{https://dl.acm.org/doi/10.1145/3460772}.

\bibitem[Mikaitis(2021)]{mikaitis_stochastic_2021}
Mantas Mikaitis.
\newblock Stochastic {Rounding}: {Algorithms} and {Hardware} {Accelerator}.
\newblock In \emph{2021 {International} {Joint} {Conference} on {Neural}
  {Networks} ({IJCNN})}, pages 1--6, July 2021.
\newblock \doi{10.1109/IJCNN52387.2021.9533756}.
\newblock URL
  \url{https://ieeexplore.ieee.org/document/9533756/?arnumber=9533756}.
\newblock ISSN: 2161-4407.

\bibitem[Bellec et~al.(2018)Bellec, Salaj, Subramoney, Legenstein, and
  Maass]{Bellec2018}
Guillaume Bellec, Darjan Salaj, Anand Subramoney, Robert Legenstein, and
  Wolfgang Maass.
\newblock Long short-term memory and learning-to-learn in networks of spiking
  neurons.
\newblock In \emph{Advances in {Neural} {Information} {Processing} {Systems}},
  volume 2018-Decem, pages 787--797, 2018.

\bibitem[Knuth(2014)]{knuth2014art}
Donald~E Knuth.
\newblock \emph{The Art of Computer Programming: Seminumerical Algorithms,
  Volume 2}.
\newblock Addison-Wesley Professional, 2014.

\bibitem[Pfeiffer and Pfeil(2018)]{pfeiffer_deep_2018}
Michael Pfeiffer and Thomas Pfeil.
\newblock Deep {Learning} {With} {Spiking} {Neurons}: {Opportunities} and
  {Challenges}.
\newblock \emph{Frontiers in Neuroscience}, 12:\penalty0 774, October 2018.
\newblock ISSN 1662-453X.
\newblock \doi{10.3389/fnins.2018.00774}.
\newblock URL
  \url{https://www.frontiersin.org/article/10.3389/fnins.2018.00774/full}.

\bibitem[Knight and Nowotny(2018)]{Knight2018}
James~C. Knight and Thomas Nowotny.
\newblock {GPUs} {Outperform} {Current} {HPC} and {Neuromorphic} {Solutions} in
  {Terms} of {Speed} and {Energy} {When} {Simulating} a {Highly}-{Connected}
  {Cortical} {Model}.
\newblock \emph{Frontiers in Neuroscience}, 12\penalty0 (December):\penalty0
  1--19, 2018.
\newblock ISSN 1662-453X.
\newblock \doi{10.3389/fnins.2018.00941}.
\newblock URL
  \url{https://www.frontiersin.org/article/10.3389/fnins.2018.00941/full}.

\bibitem[Rao et~al.(2022)Rao, Plank, Wild, and Maass]{rao_long_2022}
Arjun Rao, Philipp Plank, Andreas Wild, and Wolfgang Maass.
\newblock A {Long} {Short}-{Term} {Memory} for {AI} {Applications} in
  {Spike}-based {Neuromorphic} {Hardware}.
\newblock \emph{Nature Machine Intelligence}, 4\penalty0 (5):\penalty0
  467--479, May 2022.
\newblock ISSN 2522-5839.
\newblock \doi{10.1038/s42256-022-00480-w}.
\newblock URL \url{https://www.nature.com/articles/s42256-022-00480-w}.

\bibitem[Wunderlich and Pehle(2021)]{wunderlich_event-based_2021}
Timo~C Wunderlich and Christian Pehle.
\newblock Event-based backpropagation can compute exact gradients for spiking
  neural networks.
\newblock \emph{Scientific Reports}, 11\penalty0 (1):\penalty0 12829, December
  2021.
\newblock ISSN 2045-2322.
\newblock \doi{10.1038/s41598-021-91786-z}.
\newblock URL \url{https://doi.org/10.1038/s41598-021-91786-z}.

\bibitem[Kuon and Rose(2006)]{Kuon2006}
Ian Kuon and Jonathan Rose.
\newblock Measuring the gap between {FPGAs} and {ASICs}.
\newblock \emph{Proceedings of the internation symposium on Field programmable
  gate arrays - FPGA'06}, page~21, 2006.
\newblock \doi{10.1145/1117201.1117205}.
\newblock URL \url{http://portal.acm.org/citation.cfm?doid=1117201.1117205}.
\newblock Publisher: ACM Press Place: New York, New York, USA ISBN: 1595932925.

\end{thebibliography}

\end{document}